\RequirePackage{fix-cm}
\documentclass[twocolumn]{svjour3}          
\smartqed  
\usepackage{graphicx}
\usepackage[]{color}
\usepackage[]{amsmath}
\usepackage{amssymb}
\usepackage{textcomp}
\usepackage{subfigure}
\usepackage{hyphenat} 
\usepackage[group-separator={,}, group-minimum-digits={3}]{siunitx}

\usepackage[]{color}

\newcommand{\pf}[1]{\texttt{#1}} 
\newcommand{\units}[1]{\; \text{#1}}
\newcommand{\rev}[1]{\textcolor{black}{#1}}

\journalname{Cognitive Computation}

\begin{document}

\title{Bayesian optimisation of large-scale photonic reservoir computers\thanks{This work was supported by AFOSR (grants No. FA-9550-15-1-0279 and FA-9550-17-1-0072), the R\'egion Grand-Est, and the Volkswagen Foundation via the NeuroQNet Project.}
}

\author{Piotr Antonik \and
        Nicolas Marsal \and
        Daniel Brunner \and
        Damien Rontani 
}


\institute{P. Antonik, N. Marsal, D. Rontani \at
           LMOPS EA 4423 Laboratory \\ 
           CentraleSup\'elec \& Universit\'e de Lorraine \\
           2 rue Edouard Belin \\ 
           F-57070 Metz, France \\
           Corresponding author: P. Antonik (\email{piotr.antonik@centralesupelec.fr})           
           \and
           D. Brunner \at
           FEMTO-ST Institute/Optics Department \\
           CNRS \& Universit\'e Bourgogne Franche-Comt\'e \\
           15B Avenue des Montboucons \\
           F-25030 Besan\c{c}on, France
}

\date{Received: date / Accepted: date}

\maketitle

\begin{abstract}
  \emph{Introduction.} Reservoir computing is a growing paradigm for simplified training of recurrent neural networks, with a high potential for hardware implementations. Numerous experiments in optics and electronics yield comparable performance to digital state-of-the-art algorithms. Many of the most recent works in the field focus on large-scale photonic systems, with tens of thousands of physical nodes and arbitrary interconnections. While this trend significantly expands the potential applications of photonic reservoir computing, it also complicates the optimisation of the high number of hyper\hyp{}parameters of the system. \\
  \emph{Methods.} In this work, we propose the use of Bayesian optimisation for efficient exploration of the hyper\hyp{}parameter space in a minimum number of iteration. \\
  \emph{Results.} We test this approach on a previously reported large-scale experimental system, compare it to the commonly used grid search, and report notable improvements in performance and the number of experimental iterations required to optimise the hyper\hyp{}parameters. \\
  \emph{Conclusion.} Bayesian optimisation thus has the potential to become the standard method for tuning the hyper\hyp{}parameters in photonic reservoir computing.
\end{abstract}

\section{Introduction}

Reservoir Computing (RC) is a set of machine learning methods for designing and training artificial neural networks \cite{jaeger2004harnessing,maass2002real}. The simple idea behind this concept is to exploit the dynamics of a random recurrent neural network to process time series and only train the linear output layer by solving a (relatively simple) system of linear equations \cite{lukosevicius2009reservoir}. The reservoir computing paradigm is particularly well-suited for hardware implementations, which has attracted much interest from the community in the past ten years. The performance of the numerous experimental implementations in electronics \cite{appeltant2011information}, opto-electronics \cite{paquot2012optoelectronic,larger2012photonic,martinenghi2012photonic,larger2017high,antonik2017onlinea}, optics \cite{duport2012all,brunner2013parallel,vinckier2015high,akrout2016parallel}, and integrated on chip \cite{vandoorne2014experimental} is comparable to other digital algorithms on a series of benchmark tasks, such as wireless channel equalisation \cite{jaeger2004harnessing}, phoneme recognition \cite{triefenbach2010phoneme}, and prediction of future evolution of financial time series \cite{NFC}.

While the idea of reservoir computing greatly simplifies the training of a recurrent neural network, the optimisation of the hyper\hyp{}parameters of the network remains a full-size problem. In order to maximise performance, one has to carefully design the topology of the network, place it in the right dynamical regime (usually, task-dependent), and make sure that the scaling of the input signals is well chosen. These considerations yield a list of multiple hyper\hyp{}parameters that need to be tuned simultaneously for each benchmark task.

Most experimental implementations of reservoir computing so far share two core characteristics. First, the topology of the network, i.e. the interconnections between the different neurons, is fixed by the hardware design -- either ring-like topology for time-delay systems \cite{appeltant2011information,paquot2012optoelectronic,larger2012photonic,martinenghi2012photonic,larger2017high,duport2012all,brunner2013parallel,vinckier2015high,akrout2016parallel}, or square mesh topology for integrated systems \cite{vandoorne2014experimental,coarer2018all}. The size of the network is only fixed \emph{per se} in integrated realisations of RC, but can be modified in time-delay setups. In practice, however, it is also commonly fixed to a constant value, to avoid multiple and non-trivial re-adjustments of the experimental setup. Since the network topology and size should no longer be optimised, the list of the hyper\hyp{}parameters to tune is reduced to, typically, two variables: the input scaling, and the feedback strength. Second, most benchmark tasks used by the RC community so far typically consist of training sets of several thousands of inputs. Such (relatively) short datasets can be processed by most time-delay and integrated experiments in a matter of seconds. That is, numerous evaluations of the system performance with different values of the hyper\hyp{}parameters do not present any inconvenience. For this reason, the simple grid search has been the standard hyper\hyp{}parameter optimisation method in experimental reservoir computers so far.

A recent trend in photonic reservoir computing is the design of parallel systems to facilitate the scalability of the network and increase the processing speed. This idea has been demonstrated through frequency-multiplexing of the reservoir nodes \cite{akrout2016parallel}, and with free-space optics \cite{bueno2018reinforcement,antonik2019large,antonik2019human,dong2020optical}. 
As a result, reservoir computers of unprecedented sizes -- up to tens of thousands of physical nodes -- have been reported, and could be applied to complex tasks in computer vision, such as hand-written digit recognition \cite{antonik2019large} and human action recognition in video streams \cite{antonik2019human}. While these advances expand the potential applications of reservoir computing, they also make the optimisation of the hyper\hyp{}parameters more challenging. Similarly to the two factors above, (1) the topology of the network has to be optimised in parallel photonic experiments, which increases the number of hyper\hyp{}parameters to tune, and (2) computer vision datasets (e.g. MNIST set of handwritten digits or KTH video database of human motions) typically consist of $\num{50000} - \num{60000}$ inputs. Considering the typically low speed of certain free-space optical components, one evaluation of the experimental performance with a set of hyper\hyp{}parameters could easily take from several hours to a day. In other words, the grid search is no longer suitable for these experiments.

\rev{Consequently, more efficient methods for the optimisation of the hyper-parameters are required, such as e.g. the evolutionary-inspired genetic algorithm \cite{penkovsky2018efficient}.}
In this work, we propose the Bayesian optimisation \cite{mockus1994application,brochu2010tutorial,mockus2012bayesian,frazier2018tutorial} for this task. This idea has already been tried on time-delay \cite{yperman2016bayesian}, low-connectivity \cite{griffith2019forecasting}, \rev{and} small-dimension \cite{cerina2019lightweight} reservoir computers in numerical simulations. Here, we apply it to a more critical situation of an experimental large-scale reservoir computer, where the grid search is no longer a suitable option.
The simple idea behind Bayesian optimisation is to build a surrogate model of the cost function using Gaussian Process (GP) regression \cite{rasmussen2006gaussian}, and then efficiently sample the hyper\hyp{}parameters space by looking for regions with the most potential for improvement.

Specifically, we consider the photonic reservoir computer introduced in \cite{antonik2019human}, together with the video-based human action classification task, and apply Bayesian optimisation to tune the hyper\hyp{}parameters of the experiment. We perform numerical and experimental investigations, and report notable performance improvements in both cases. Furthermore, Bayesian optimisation offers a better understanding of the importance of different hyper\hyp{}parameters, i.e. it helps to differentiate the significant parameters from those that have little impact on the system performance. Considering the above advantages, Bayesian optimisation could become the standard hyper\hyp{}parameters optimisation method for large-scale photonic reservoir computers.

\section{Methods}

We start by briefly reviewing the basic principles of reservoir computing (Sec. \ref{subsec:rc}). Then, we present the experimental reservoir computer and its hyper\hyp{}parameters (Sec. \ref{subsec:exp}) and the human actions classification task (Sec. \ref{subsec:kth}), originally introduced in \cite{antonik2019human}, used here to test the Bayesian optimisation approach, presented in Sec. \ref{subsec:bayes}.

\subsection{Basic principles of reservoir computing}
\label{subsec:rc}

A typical discrete-time reservoir computer contains a large number $N$ of internal variables $x_{i\in 0 \ldots N-1} (n)$ evolving in discrete time $n \in \mathbb{Z}$, as given by
\begin{equation}
  x_i(n+1) = f_\text{nl} \left( \sum_{j=0}^{N-1} W_{ij} x_j(n) + \sum_{j=0}^{K-1} b_{ij} u_j(n) \right).
  \label{eq:rcevo2}
\end{equation}
where $f_\text{nl}$ is the nonlinear function, $u_j(n)$ is the input signal of dimension $K$, $b_{ij}$ is the $N \times K$ matrix of input weights, often referred to as the ``input mask'', and $W_{ij}$ is the $N\times N$ matrix of interconnecting weights between the neurons of the neural network. 

The reservoir computer produces an output signal $y(n)$, given by a linear combination of the states of its internal variables
\begin{equation}
  y(n) = \sum_{i=0}^{N-1} w_i x_i (n),
  \label{eq:rcout}
\end{equation}
where $w_i$ are the readout weights, trained either offline (using standard linear regression methods, such as the ridge regression algorithm \cite{tikhonov1995numerical} used here), or online \cite{antonik2017online}, in order to minimise the Normalised Mean Square Error (NMSE) between the output signal $y(n)$ and the target signal $d(n)$, given by
\begin{equation}
  \text{NMSE} = \frac{\left\langle \left( y(n) - d(n) \right)^2 \right\rangle}{\left\langle \left( d(n) - \langle d(n) \rangle \right)^2 \right\rangle} .
  \label{eq:nmse}
\end{equation}

\subsection{Photonic reservoir computer and its hyper\hyp{}parameters}
\label{subsec:exp}

\begin{figure}[t]
  \centering
  \includegraphics[width=0.45\textwidth]{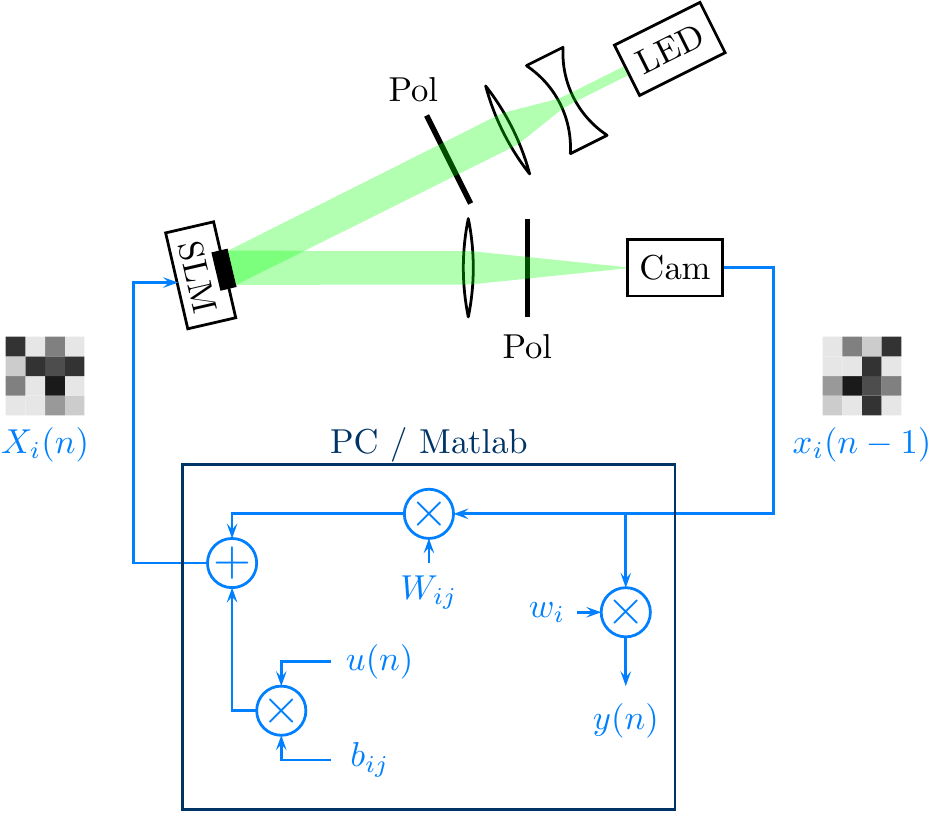}
\caption{Scheme of the experimental setup, composed of an optical arm (top half) and digital electronics (bottom half, rendered in blue). The optical part is composed of a light source (green LED), a pair of lenses to expand the beam to match the surface of the SLM, and a linear polariser rotated accordingly to the fast axis of the SLM. The spatial light modulator is imaged by a camera through an imaging lens and a second polariser, that transforms the phase modulation into intensity modulation. The electronics part is composed of a computer, running Matlab, that captures the reservoir states $x_i$ from the camera, evaluates the outputs $y(n)$, computes the inputs $X_i$ to the SLM and loads them to the device.}
  \label{fig:exp}
\end{figure}

The experimental setup, introduced in \cite{antonik2019human}, is schematised in Fig. \ref{fig:exp}. 
It is composed of a free-space optical arm and digital electronics, depicted in blue. 
The optical beam, generated by a green LED source at $530\units{nm}$ (Thorlabs M530L3), is linearly polarised, collimated, and expanded to roughly $17 \units{mm}$ in diameter to evenly lighten the $7.68 \units{mm} \times 7.68 \units{mm}$ surface of the spatial light modulator (Meadowlark XY Phase P512 -- 0532). 
The SLM is imaged by a high-speed camera (Allied Vision Mako U-130B) after focusing the light beam with the imaging lens and transforming phase modulation (induced by the liquid crystals of the SLM) into intensity modulation through a second polariser.

The experimental setup implements the nonlinear function $f_\text{nl}$ in Eq. \ref{eq:rcevo2}, \rev{that can be modelled as 
\begin{equation}
  f_\text{nl} ( X_i(n) ) = \left\lfloor I_0 \sin^2 \left( \left\lfloor X_i(n) \right\rfloor_8 \right) \right\rfloor_{10}
  \label{eq:nlfun}
\end{equation}
where $X_i(n)$ is the argument of the function, defined below, $I_0$ is the intensity of the illuminating beam and $\lfloor \rfloor_{8,10}$ are the 8-bit and 10-bit quantifications due to the SLM and the camera, respectively \cite{antonik2019large,antonik2019human}.}
The rest of the Eq. \ref{eq:rcevo2} is computed in Matlab. 
At each discrete timestep $n$, the input to the nonlinear function 
\begin{equation}
  X_i (n) = \sum_{j=0}^{N-1} W_{ij} x_j(n) + \sum_{j=0}^{K-1} b_{ij} u(n)
  \label{eq:pref}
\end{equation}
is computed, and the resulting matrix is loaded onto the SLM device. 
\rev{The polarisation-filtered SLM image, formed via an imaging lense, is recorded by the camera.}
The resulting data corresponds to the reservoir state $x_i(n+1) = f_\text{nl}( X_i (n) )$.

The input mask $b_{ij}$ and the interconnection matrix $W_{ij}$ are generated randomly in the beginning of the experiment. The input mask $b_{ij}$ is initially drawn from a uniform distribution over the interval $[-1, +1]$ as in \cite{rodan2011minimum,paquot2012optoelectronic,duport2012all}, and then multiplied by a global scaling factor $\beta$, called the \emph{input gain}. The interconnection matrix $W_{ij}$ is generated as follows. First, a diagonal matrix of size $N\times N$ is created and multiplied by a coefficient $\alpha$, called the \emph{feedback gain}, since the diagonal elements of $W_{ij}$ are responsible for the feedback of the reservoir, i.e. the connection of each neuron to its past states. A fraction $\rho$ of the off-diagonal elements of $W_{ij}$ are assigned a fixed value $\gamma$, while all the others are set to zero. The off-diagonal elements correspond to the connections between different neurons. Therefore, the connectivity of the network is defined by two parameters: the \emph{interconnection density} $\rho$ and the \emph{interconnection gain} $\gamma$.
In summary, the dynamics of the reservoir computer depend on four hyper\hyp{}parameters: $\alpha, \beta, \gamma,$ and $\rho$, recapped in Tab. \ref{tab:hyperparams}.

\rev{The processing speed of the system depends on two main factors: (i) the time Matlab requires to compute the next SLM matrix (which increases with the reservoir size) and (ii) the communication speed between Matlab and the SLM (which is independent of the reservoir size).
  The experiment is capable of processing 7 video frames per second with the smallest reservoir ($N=\num{1024}$) and 2 frames per second with the largest reservoir ($N=\num{16384}$).
Therefore, the total classification of the KTH database with roughly \num{53000} inputs (see Sec. \ref{subsec:kth}) takes from approximately 2 to 7 hours.}

\subsection{\rev{Bayesian optimisation of hyper\hyp{}parameters}}
\label{subsec:bayes}

Many optimisation problems in machine learning are ``black-box'' problems, where the objective function $F(x)$ is unknown. 
In our study, $F(x)$ is the performance of the reservoir computer on the KTH dataset (see Sec. \ref{subsec:kth}), i.e. the classification accuracy, as function of the four hyper\hyp{}parameters (see Sec. \ref{subsec:exp}): $ \pf{accuracy} = F(\alpha, \beta, \gamma, \rho)$. 
Finding the maximal value of $F(\alpha, \beta, \gamma, \rho)$ is a key step in obtaining the optimal performance from the reservoir computer.

If $F$ is computationally cheap to evaluate, one can sample the hyper\hyp{}parameter space at many points e.g. via grid or random search. 
Grid search has been the standard approach in the photonic reservoir computing field so far \cite{appeltant2011information,paquot2012optoelectronic,larger2012photonic,martinenghi2012photonic,larger2017high,duport2012all,brunner2013parallel,vinckier2015high,akrout2016parallel,sande2017advances}.
However, if the function evaluation is expensive -- such as in our case, where one experimental evaluation of $F(\alpha, \beta, \gamma, \rho)$ takes from \rev{2 to 7 hours (see Sec. \ref{subsec:exp})} -- it is important to minimise the number of samples of $F$ required to find its optimum.

The Bayesian optimisation technique attempts to find the global optimum in a minimum number of steps. It is an iterative approach that builds a surrogate model to approximate the objective function $F$, the former being much cheaper to evaluate. The sampling of new points in the hyper\hyp{}parameter space is guided by an acquisition function, which estimates the hyper\hyp{}parameter regions of most uncertainty and most gain, i.e. where an improvement over the current best observation (optimum) is the most likely.

A popular surrogate model for Bayesian optimisation is the GP regression \cite{rasmussen2006gaussian}, a non-parametric kernel-based probabilistic model. 
Unlike linear regression, which seeks the best parameters that fit a linear model onto data, the GP approach finds a distribution over the possible functions $f(x)$ that are consistent with the observed data. 
The Bayesian approach consists in building a starting GP model of $F(\alpha, \beta, \gamma, \rho)$ from an initial set of observations, and then updating the model as new data points are being observed.
An observation, in this context, is the performance of the RC (the accuracy) for a certain combination of hyper\hyp{}parameters $(\alpha, \beta, \gamma, \rho)$.
The set of possible functions $f(x)$ for the GP model is defined by specifying their smoothness. This is achieved through a covariance or kernel function, of the model. 
In practice, choosing a kernel function often requires an initial guess by the user, while approaches exist to optimising the kernel through cross-validation \cite{mackay1992practical}.

The acquisition function is executed over the GP prediction of the objective function $F(\alpha, \beta, \gamma, \rho)$. 
It takes into account the GP model with its uncertainties to evaluate the potential improvement over the current optimum in each point of the hyper\hyp{}parameters space. 
Intuitively, it provides a trade-off between \emph{exploitation} of the region close to the current optimum -- i.e. observation of the neighbour points -- and \emph{exploration} -- the probing of different regions of the hyper\hyp{}parameters space in search for another possible optimum.
In this work, we use the \emph{expected improvement} function, that evaluates the expected amount of improvement in the objective function, ignoring values that cause an increase in the objective (see \cite{brochu2010tutorial} for a review of several acquisition functions).

The Bayesian optimisation of our reservoir computer can be summarised as follows:
\begin{enumerate}
  \item Run the reservoir computer (numerically or experimentally) several times (typically 5-10) to build a starting set of observations. 
  \item Compute the GP model from the observation set.
  \item Evaluate the acquisition function over the entire hyper\hyp{}parameters space and find its maximum, which becomes the candidate for the next observation.
  \item Set the new values of the hyper\hyp{}parameters and run the reservoir computer. Add the resulting observation to the set.
  \item Repeat steps 2 to 4 until the desired accuracy has been achieved.
\end{enumerate}

The Bayesian optimisation is supported by Matlab and can be implemented using the provided functions. The GP regression is carried out by the \pf{fitrgp} function. 
\rev{To avoid influencing the algorithm with any \emph{a priori} knowledge we might possess, we let it choose its optimal parameters (such as the basis function, the kernel function and its parameters) through cross-validation by setting the option \pf{OptimizeHyperparameters} to \pf{all} \cite{matlab_fitrgp}.}
For the acquisition function, we chose the \pf{expected\hyp{}improvement} function, readily implemented in Matlab \cite{matlab_bayes} and defined by:
\begin{equation}
  \text{EI} (x) = \max \left( 0, F_\text{best} - F(x) \right)
  \label{eq:ei}
\end{equation}
where $F_\text{best}$ is the optimal value of $F(\alpha, \beta, \gamma, \rho)$ observed so far. On top of the standard Bayesian optimisation procedure described above, we added a special modification to check that every new observation candidate is a previously unprobed, new set of parameters. This allows the algorithm to scan the hyper\hyp{}parameters space faster by avoiding the repetition of identical parameter values.

\begin{figure*}[t]
  \centering
  \subfigure[2 observations]{\includegraphics[width=0.24\textwidth]{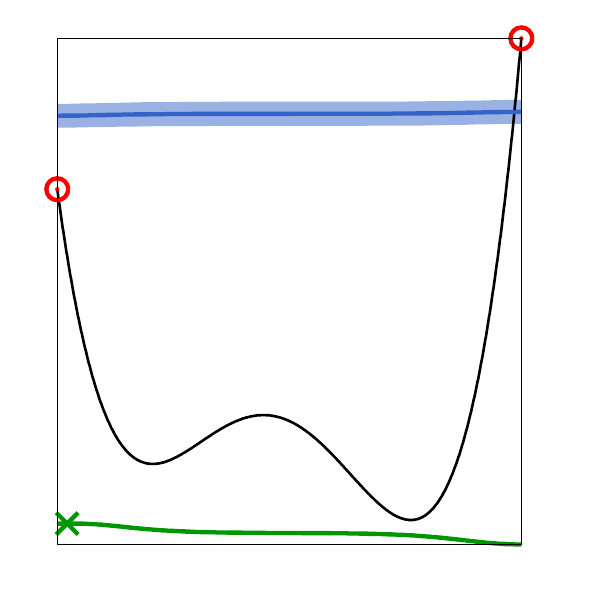}\label{subfig:toy1}}
  \subfigure[5 observations]{\includegraphics[width=0.24\textwidth]{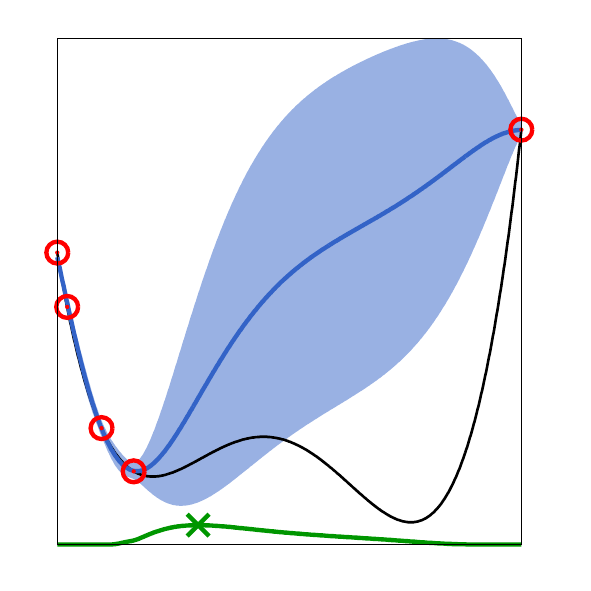}\label{subfig:toy2}}
  \subfigure[9 observations]{\includegraphics[width=0.24\textwidth]{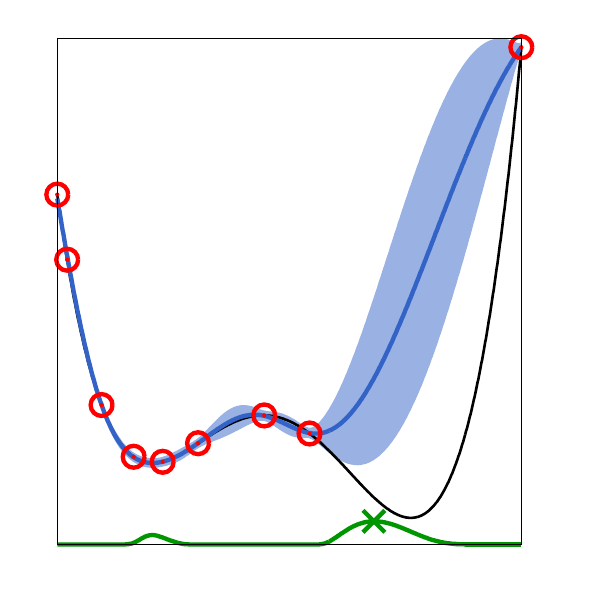}\label{subfig:toy3}}
  \subfigure[12 observations]{\includegraphics[width=0.24\textwidth]{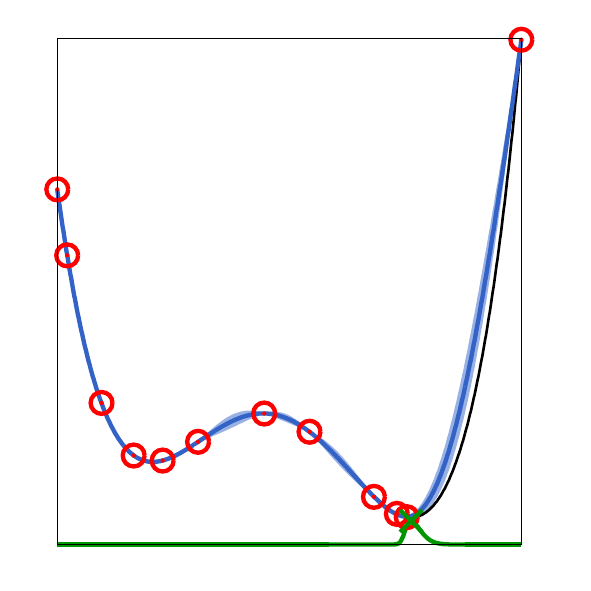}\label{subfig:toy4}}
  \caption{Illustration of the Bayesian optimisation on a toy 1D problem. The target function is shown in black, the red markers correspond to the observations, the acquisition function is rendered in green, and the GP model is shown in blue (the shade corresponds to the uncertainty).}
  \label{fig:toy}
\end{figure*}

Figure \ref{fig:toy} illustrates the Bayesian optimisation algorithm in action on a toy example in one dimension. 
The objective function, displayed in black, was chosen to present one local and one global minimum. 
Red markers show the observations of the target function, the acquisition function is shown in green, and the GP model is shown in blue with shaded uncertainty. 
The green markers indicate the acquisition function's maxima, which are the candidates for the following observations.
For visual clarity, the plot of the acquisition function was rescaled at each step.
Figure \ref{subfig:toy1} shows the stage after the two starting observations, that were used to initialise the GP model. Figure \ref{subfig:toy2} shows that the model has located the local minimum region after three additional observations, with a significant uncertainty in the right-hand region. In Fig. \ref{subfig:toy3}, after another four observations, the acquisition function forces the process away from the local minimum on the left in order to explore the right-hand side region. Finally, in Fig \ref{subfig:toy4}, after a total of 12 observations, the model has found another minimum region. The low overall uncertainty indicates that it is the global minimum of the function, that can now be exploited to pin-point the exact optimal value.

\subsection{\rev{KTH human actions classification task}}
\label{subsec:kth}

\begin{figure}[t]
\centering
\includegraphics[width=0.15\textwidth]{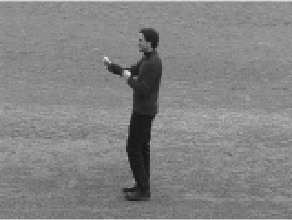}
\includegraphics[width=0.15\textwidth]{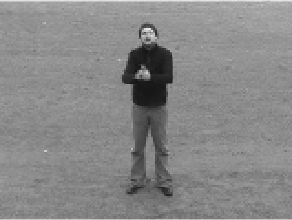}
\includegraphics[width=0.15\textwidth]{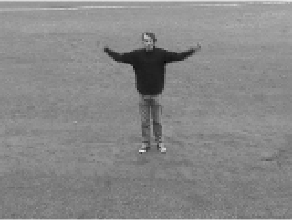}
\includegraphics[width=0.15\textwidth]{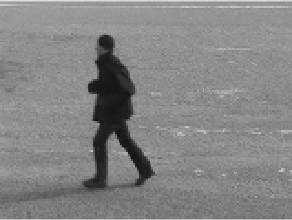}
\includegraphics[width=0.15\textwidth]{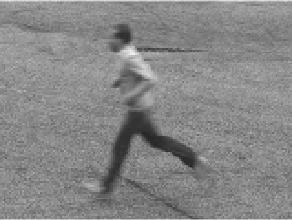}
\includegraphics[width=0.15\textwidth]{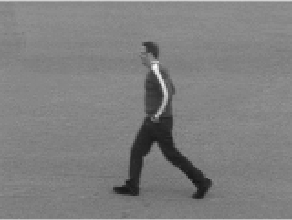}
\caption{Examples of action frames from the KTH database, from left to right: boxing, hand clapping, hand waving, jogging, running, and walking. Six different subjects are illustrated out of the total of 25. All videos have been taken outdoors over a homogeneous background, which corresponds to the ``s1'' subset of the full database.}
\label{fig:kth}
\end{figure}

Similarly to Ref. \cite{antonik2019human}, we used the KTH database of human actions \cite{schuldt2004recognizing}, limited to the first ``s1'' scenario.
The video database contains six types of human actions -- walking, jogging, running, boxing, hand waving, and hand clapping (illustrated in Fig. \ref{fig:kth}) -- performed 4 times by 25 subjects, for a total of $600$ video sequences.
They vary in length and contain between $24$ and $239$ frames.
All videos were recorded over homogeneous background with a static camera at $25 \units{fps}$ and downsampled to the spatial resolution of $160\times120$ pixels. 

We used the Histograms of Gradients (HOG) algorithm \cite{dalal2005histograms,bahi2015robust} to extract the relevant features from the video frames.
The main idea of this technique is that local object appearance and shape can often be expressed well enough by distribution of local intensity gradients or edges' directions \cite{antonik2019human}.
The computation of HOG features was performed in Matlab, individually for each frame of every sequence using the built-in \pf{extractHOGFeatures} function with a cell size of $8\times 8$ and a block size of $2\times 2$. Given the frame size of $160 \times 120$ pixels, we obtained $\num{9576}$ features per frame. 
We then applied the principal component analysis (PCA) \cite{pearson1901liii.,hotelling1933analysis} based on the covariance method \cite{smith2002tutorial}, to reduce the number of features down to \num{2000}, keeping $91.6\%$ of total variance.

The reservoir computer was trained over a subset of 450 video sequences, and tested over the remaining 150 sequences. We trained 6 binary classifiers (i.e. the output nodes), each for one motion class, and applied the winner-takes-all approach between them. The final decision over the duration of a sequence was made by taking the most frequent class in the reservoir output.
\rev{The classification accuracy is defined as the ratio of the correctly recognised video sequences in the testing set over the total number of 150 sequences.}

\section{Results}
\label{sec:res}

\subsection{Performance improvement}
\label{subsec:perf}

To demonstrate the performance improvement offered by the Bayesian optimisation, in comparison to the simple grid search, we employ both methods on the same photonic reservoir computer, described in Sec. \ref{subsec:exp}, and applied to the same benchmark task, presented in Sec. \ref{subsec:kth}. Furthermore, we evaluate the system's performance both in experiments and in numerical simulations. Similarly to the original work on this experiment \cite{antonik2019human}, we consider reservoir sizes from $\num{1024}$ to $\num{16384}$ nodes. 

The optimisation of our four hyper\hyp{}parameters without \textit{a priori} knowledge of the regions of better performance is a non-trivial task for grid search, especially when the evaluation of a set of values can take up to several hours experimentally. In order to keep manageable experimental times, we had to severely restrict the grid search intervals down to 2-3 values for each parameters. This approach allows to determine the right scaling of each parameter, but lacks the resolution to find the most optimal values. Table \ref{tab:hyperparams} contains the allowed values for each parameter used in the grid search.

\begin{table*}
  \centering
  \begin{tabular}{l|c|c|c}
    Parameter & Symbol & Search values (grid search) & Search intervals (Bayesian optimisation) \\
    \hline
    Feedback gain & $\alpha$ & $0.6, 0.8, 1.0$ & $0.1 - 1.5$ \\
    Input gain & $\beta$ & $0.01, 0.1$ & $10^{-10} - 1$ \\
    Interconnectivity gain & $\gamma$ & $0.001, 0.01, 0.1$ & $10^{-10} - 1$ \\
    Interconnectivity density & $\rho$ &  $0.001, 0.01, 0.1$ & $10^{-10} - 1$ \\
  \end{tabular}
  \caption{Hyper\hyp{}parameters search intervals for the two optimisation approaches.}
  \label{tab:hyperparams}
\end{table*}

The Bayesian optimisation is only affected by the dimensionality of the hyper\hyp{}parameter space in the sense that it requires a larger starting set of observations to fit an accurate enough GP model onto the data and start sampling the right regions for improvement. Our trials have shown that 8 starting observations were enough to properly initialise the GP model. To truly test the Bayesian optimisation approach, we used a much larger, and fine-grained hyper\hyp{}parameter space than with the grid search. Moreover, despite gaining some intuition on the optimal parameters with the grid search, we made sure not to disclose any \textit{a priori} knowledge to the GP model. That is, the starting observations were chosen from the extreme values in the hyper\hyp{}parameter space, with one observation in the middle to force a non-linear fit. Table \ref{tab:hyperparams} shows the intervals used for the Bayesian optimisation.

\begin{figure*}[t]
  \centering
  \includegraphics[width=0.6\textwidth]{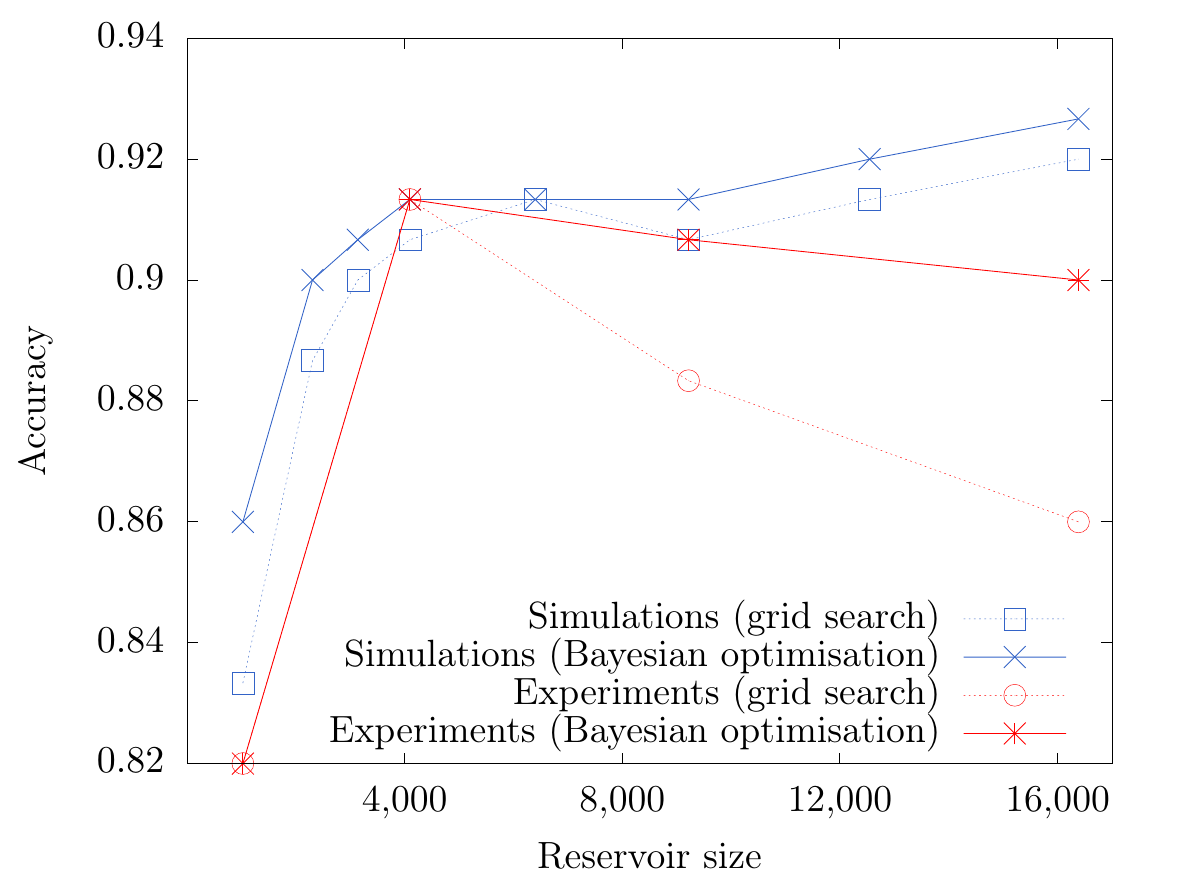}
  \caption{Performance of numerical (blue lines) and experimental (red lines) reservoir computers of different sizes, optimised with either the grid search (dotted lines and hollow markers) or the Bayesian approach (solid lines and cross markers), with search intervals and values given in Tab. \ref{tab:hyperparams}. In most cases, the Bayesian optimisation outperforms the grid search (and matches in the worst ones), with the accuracy increase of up to $4\%$ (in the case of the largest experimental reservoir). This is a significant improvement in the field of classification tasks, where the last fractions of percent are the hardest to gain.}
  \label{fig:res}
\end{figure*}

Figure \ref{fig:res} shows the results obtained with both optimisation methods, experimentally (in red) and numerically (in green), with reservoirs of different sizes. Each point corresponds to the highest accuracy we could obtain with the corresponding approach. In each experiment, the Bayesian optimisation (plus markers and solid lines) either matches, or outperforms the grid search (round markers and dotted lines). The improvement is quite significant in numerical simulations with a small ($N=\num{1024}$) reservoir -- from $83.3\%$ to $86.0\%$, and in experiments with a large ($N=\num{16384}$) reservoir -- from $86.0\%$ to $90.0\%$. Comparing the accuracy of the present setup ($91.3\%$, see \cite{antonik2019human}) to the state-of-the-art result of $95.6\%$ reported in \cite{shi2015learning}, a $4\%$ increase in performance is a valuable improvement.

\begin{figure*}[t]
  \centering
  \subfigure[5 observations]{\includegraphics[width=0.45\textwidth]{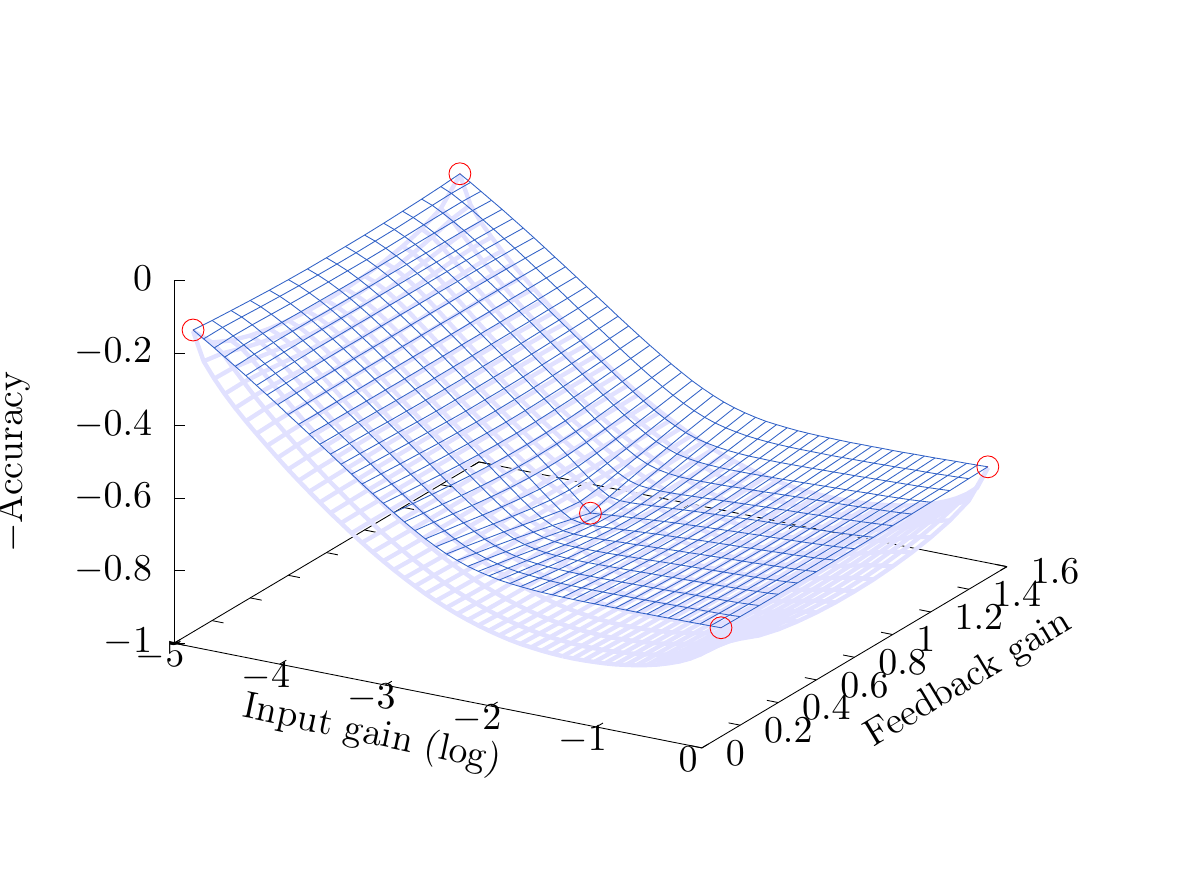}\label{subfig:demo1}}
  \subfigure[10 observations]{\includegraphics[width=0.45\textwidth]{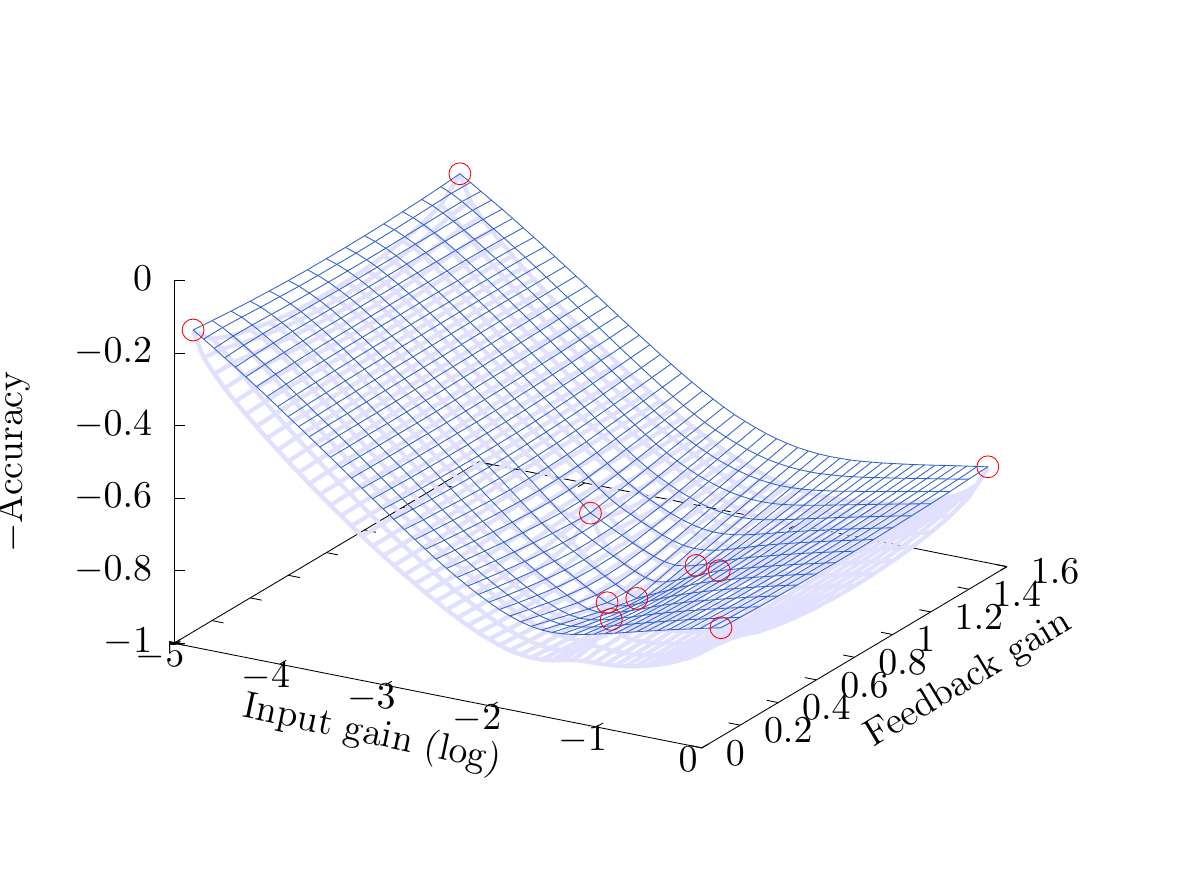}\label{subfig:demo2}}
  \subfigure[14 observations]{\includegraphics[width=0.45\textwidth]{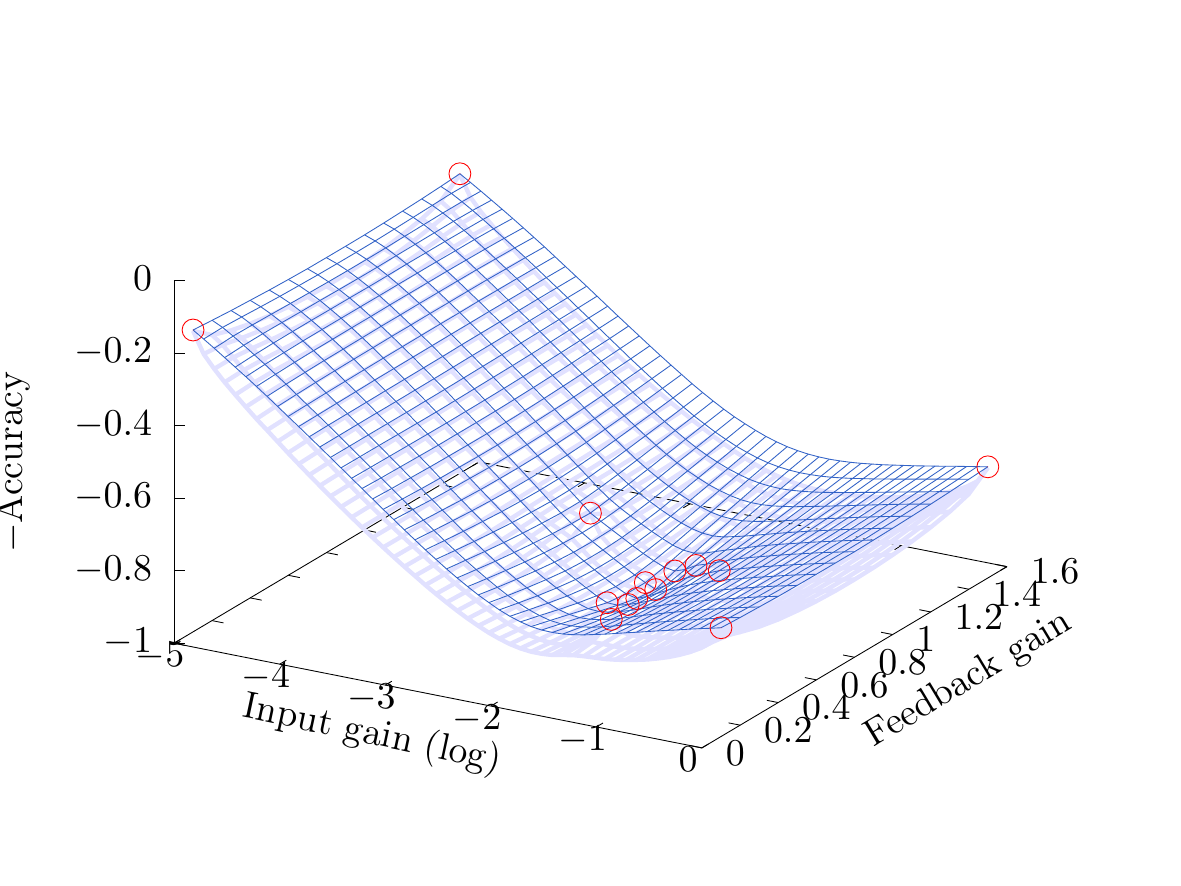}\label{subfig:demo3}}
  \subfigure[19 observations]{\includegraphics[width=0.45\textwidth]{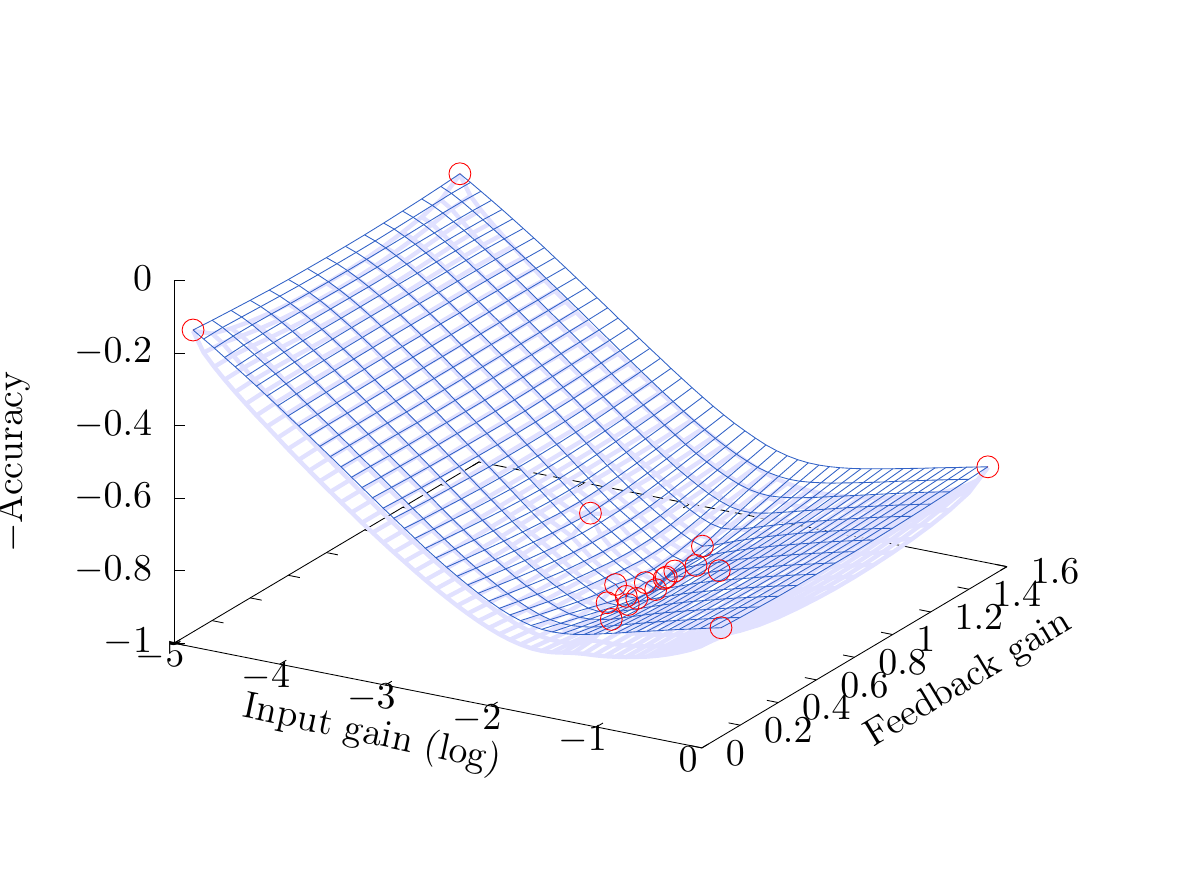}\label{subfig:demo4}}
  \caption{Illustration of the Bayesian optimisation on a small reservoir computer ($N=\num{1024}$) with two hyper\hyp{}parameters (input and feedback gains). The interconnection density and gain are fixed (i.e. not optimised in this example) for the sake of simplicity. The accuracy is plotted with a negative sign for a better visualisation. \textbf{(a)} The starting set of 5 observations (red marks), the fitted GP model (blue) and its lower uncertainty (light blue). Upper uncertainty is omitted, again, for the sake of clarity. \textbf{(b)} The process starts with the exploration of the flat region of the model with the lowest uncertainty. \textbf{(c)} The process discovers a pit in the cost function towards the lower values of the feedback gain and exploits it to find the optimum. \textbf{(d)} The optimum is found after 19 iterations, with the best accuracy of $86\%$ (see Fig. \ref{fig:res}).}
  \label{fig:demo}
\end{figure*}

Figure \ref{fig:demo} illustrates the functioning of the Bayesian optimisation process on the hyper\hyp{}parameters of the reservoir computer with $N=\num{1024}$ nodes. For the sake of visualisation, we fixed two parameters to their optimal values -- the interconnection density $\rho$ and the interconnection gain $\gamma$ -- and ran the optimisation of the two remaining hyper\hyp{}parameters: the feedback gain $\alpha$ and the input gain $\beta$, so that the results could be plotted on a 3D graph. After collecting 5 observations of the cost function, we fit a GP model and evaluate the acquisition function (not displayed here) to guide the sampling of the hyper\hyp{}parameter space (Fig. \ref{subfig:demo1}). The geometry of the GP model's cost function -- a rapidly rising fraction on the left-hand side and a moderately flat fraction on the right-hand side -- suggests the exploration of the flat region with the most promising uncertainty. After 5 additional observations there the process discovers a pit (Fig. \ref{subfig:demo2}) and starts exploiting (Fig. \ref{subfig:demo3}). The optimal accuracy of $86\%$ (as indicated in Fig. \ref{fig:res}) is found after 19 iterations of the algorithm (Fig. \ref{subfig:demo4}). 
\rev{As a side note, one should keep in mind that the shape of the objective function (i.e. the geometry of the GP model) highly depends on the task at hand (here, the classification of videos from the KTH dataset) and the definition of the accuracy (see Sec. \ref{subsec:kth}).}

\subsection{Time gain}
\label{subsec:time}

The Bayesian approach not only improves the RC performance, but also requires less time to find the optimal parameters, i.e. minimises the number of time-consuming iterations of the experiment. 
Without \textit{a priori} intuition on the optimal settings, the grid search needs to be executed over all 54 combinations of the hyper\hyp{}parameters values from Tab. \ref{tab:hyperparams} in order to find the best composition. 
\rev{Bayesian optimisation, on the other hand, requires between 19 iterations (for a small reservoir with $N=\num{1024}$ nodes) and 39 iterations (for a large reservoir with $N=\num{16384}$ nodes), which corresponds to a $65\%-28\%$ time gain, respectively.}

\rev{Another} significant advantage of an iterative algorithm over the grid search is that it autonomously converges towards the optimal value, and could, if necessary, be stopped\rev{, e.g. when the desired accuracy has been achieved.}
To illustrate this idea, consider the experimental results with the largest reservoir of $\num{16384}$ nodes. 
The Bayesian approach managed to find the optimum after 39 iterations (including the starting observations), which represents roughly a 30\% gain in time. 
Interestingly, a slightly less optimal combination of hyper\hyp{}parameters, with an accuracy loss of $1.3\%$, was found after only 22 iterations. 
That is, a speed gain of roughly 60\% can be obtained with no more than $1.3\%$ loss in performance should time be a more restricted commodity.
Such flexibility cannot be achieved with grid search.

\subsection{Structure of the hyper\hyp{}parameters space}
\label{subsec:shape}

Another indirect result provided by Bayesian optimisation is a better understanding of the relative importance of the four hyper\hyp{}parameters $\alpha$, $\beta$, $\gamma$, and $\rho$. Grid search only evaluates the combinations that the user believes to be relevant, while Bayesian optimisation uses the acquisition function (see Sec. \ref{subsec:bayes}) to explore the entire hyper\hyp{}parameters space and in particular the regions of high uncertainty, where an improvement could potentially be found. Therefore, by exploring the regions a user might not have thought of, it provide additional insights about the cost function's shape and the relative impact of the different dimensions.

Based on this approach we learned through numerical simulations that the system is only sensitive to the first two parameters -- the input scaling $\beta$ and the feedback gain $\alpha$. In other words, those parameters need to be accurately adjusted to maximise the accuracy, while the remaining two parameters -- the interconnection gain $\gamma$ and the interconnection density $\rho$ -- can take multiple (very) different values within the ranges we studied. To illustrate this result, we ran the Bayesian optimisation of a small numerical reservoir computer with $N=\num{1024}$ nodes for 500 iterations, to let it explore the hyper\hyp{}parameters space. Out of 500 observations, the maximum accuracy of $86\%$ was obtained in 116 different points. However, these points only differ in the values of $\gamma$ and $\rho$, which take values within $[10^{-10}, 10^{-5.5}]$ and $[10^{-10}, 1]$, respectively, while the first two parameters take the same exact values of $\beta=0.0158$ and $\alpha=1$. 

\rev{The same outcome appears with different reservoir sizes, as summarised in Tab. \ref{tab:hypervalues}.
The first two parameters, $\alpha$ and $\beta$, only slightly vary with different reservoir sizes, without a noticeable trend. In general, the system works best with a high feedback gain $\alpha$ (memory capacity is required to store the information from previous frames) and a relatively low input scaling $\beta$, which is mostly due to the large dimensionality of the input signal.
As for $\gamma$ and $\rho$, optimal performance can be obtained in multiple points of the ($\gamma,\rho$)-plane.
In other words, reservoirs with different topologies, i.e. networks with different interconnection matrices, perform the classification equally as good. On the other hand, the spectral radii of these matrices remain very similar, which is logical, since the spectral radius determines the dynamics of the system and how it processes the input information.}
These findings call for an in-depth study of the properties of the interconnection matrix $W_{ij}$ in large-scale photonic reservoir computers, that we leave for future work.

\begin{table*}
  \centering
  \begin{tabular}{c|c|c|c|c||c|c|c|c}
    & \multicolumn{4}{c}{Grid search} & \multicolumn{4}{c}{Bayesian optimisation} \\
    Reservoir size & $\alpha$ & $\beta$ & $\log\gamma$ & $\log\rho$ & $\alpha$ & $\beta$ & $\log \gamma$ & $\log \rho$ \\
    \hline
    \num{1024} & $0.8$ & $0.01$ & $-1$ & $-2$ & 1 & $0.0158$ & $[-5.5, -10]$ & $[0, -10]$ \\
    \num{4096} & $0.6$ & $0.1$ & $-2$ & $-3$ & 0.9 & $0.0398$ & $-4.8$ & $-0.1$ \\
    \num{6400} & $0.6$ & $0.1$ & $-1$ & $-2$ & 1.0 & $0.0501$ & $[-6.8, -7.3]$ & $[-4.8, -5.6]$ \\
    \num{9216} & $0.8$ & $0.01$ & $-1$ & $-2$ & 0.8 & $0.0316$ & $[-0.4, -10]$ & $[0, -10]$ \\
    \num{12544} & $1.0$ & $0.01$ & $-3$ & $-2$ & 0.6 & $0.01$ & $[0, -10]$ & $[0, -10]$ \\
    \num{16384} & $1.0$ & $0.1$ & $-3$ & $-1$ & 0.9 & $0.0079$ & $[0, -10]$ & $[0, -10]$ \\
  \end{tabular}
  \caption{\rev{Optimal values of hyper\hyp{}parameters obtained in numerical simulations for different reservoir sizes.}}
  \label{tab:hypervalues}
\end{table*}

\section{Conclusion}
\label{sec:ccl}

In this work, we proposed the Bayesian optimisation algorithm for tuning the hyper\hyp{}parameters in large-scale photonic reservoir computers. We tested this approach on a previously reported experimental system, applied to a challenging task in computer vision, and compared the results to the grid search, commonly used by the RC community. We report improvements in terms of (1) the classification performance, with an accuracy increase up to $4\%$, and (2) the convergence time to the optimal set of hyper\hyp{}parameters, with a roughly $30\%$ gain in time (that could be doubled for a less than $1.5\%$ accuracy penalty). Taking into account the proximity of the accuracy of our photonic reservoir computer to the state-of-the-art results on this task, and the experimental hyper\hyp{}parameters optimisation time measured in days, these improvements prove to be precious enhancements of the system performance. Furthermore, extensive exploration of the hyper\hyp{}parameters space with the Bayesian method offers valuable insights on its underlying structure and the relative importance of the parameters. Considering all the advantages offered by the Bayesian optimisation algorithm, it may soon become the new standard approach for the optimisation of hyper\hyp{}parameters in photonic reservoir computing.


%
\section*{Compliance with Ethical Standards}

\noindent
\textbf{Conflict of Interest.} The authors declare that they have no conflict of interest.

\noindent
\textbf{Ethical approval.} This article does not contain any studies with human participants or animals performed by any of the authors.

%

\bibliographystyle{spphys}       
\bibliography{refs.bib}

\begin{thebibliography}{10}
\providecommand{\url}[1]{{#1}}
\providecommand{\urlprefix}{URL }
\expandafter\ifx\csname urlstyle\endcsname\relax
  \providecommand{\doi}[1]{DOI \discretionary{}{}{}#1}\else
  \providecommand{\doi}{DOI \discretionary{}{}{}\begingroup
  \urlstyle{rm}\Url}\fi

\bibitem{jaeger2004harnessing}
H.~Jaeger, Science \textbf{304}(5667), 78 (2004).
\newblock \doi{10.1126/science.1091277}

\bibitem{maass2002real}
W.~Maass, T.~Natschläger, H.~Markram, Neural Computation \textbf{14}(11), 2531
  (2002).
\newblock \doi{10.1162/089976602760407955}

\bibitem{lukosevicius2009reservoir}
M.~Luko{\v{s}}evi{\v{c}}ius, H.~Jaeger, Computer Science Review \textbf{3}(3),
  127 (2009).
\newblock \doi{10.1016/j.cosrev.2009.03.005}

\bibitem{appeltant2011information}
L.~Appeltant, M.~Soriano, G.V. der Sande, J.~Danckaert, S.~Massar, J.~Dambre,
  B.~Schrauwen, C.~Mirasso, I.~Fischer, Nature Communications \textbf{2}(1)
  (2011).
\newblock \doi{10.1038/ncomms1476}

\bibitem{paquot2012optoelectronic}
Y.~Paquot, F.~Duport, A.~Smerieri, J.~Dambre, B.~Schrauwen, M.~Haelterman,
  S.~Massar, Scientific Reports \textbf{2}(1) (2012).
\newblock \doi{10.1038/srep00287}

\bibitem{larger2012photonic}
L.~Larger, M.C. Soriano, D.~Brunner, L.~Appeltant, J.M. Gutierrez, L.~Pesquera,
  C.R. Mirasso, I.~Fischer, Optics Express \textbf{20}(3), 3241 (2012).
\newblock \doi{10.1364/oe.20.003241}

\bibitem{martinenghi2012photonic}
R.~Martinenghi, S.~Rybalko, M.~Jacquot, Y.K. Chembo, L.~Larger, Physical Review
  Letters \textbf{108}(24) (2012).
\newblock \doi{10.1103/physrevlett.108.244101}

\bibitem{larger2017high}
L.~Larger, A.~Bayl{\'{o}}n-Fuentes, R.~Martinenghi, V.S. Udaltsov, Y.K. Chembo,
  M.~Jacquot, Physical Review X \textbf{7}(1) (2017).
\newblock \doi{10.1103/physrevx.7.011015}

\bibitem{antonik2017onlinea}
P.~Antonik, M.~Haelterman, S.~Massar, Cognitive Computation \textbf{9}(3), 297
  (2017).
\newblock \doi{10.1007/s12559-017-9459-3}

\bibitem{duport2012all}
F.~Duport, B.~Schneider, A.~Smerieri, M.~Haelterman, S.~Massar, Optics Express
  \textbf{20}(20), 22783 (2012).
\newblock \doi{10.1364/oe.20.022783}

\bibitem{brunner2013parallel}
D.~Brunner, M.C. Soriano, C.R. Mirasso, I.~Fischer, Nature Communications
  \textbf{4}(1) (2013).
\newblock \doi{10.1038/ncomms2368}

\bibitem{vinckier2015high}
Q.~Vinckier, F.~Duport, A.~Smerieri, K.~Vandoorne, P.~Bienstman, M.~Haelterman,
  S.~Massar, Optica \textbf{2}(5), 438 (2015).
\newblock \doi{10.1364/optica.2.000438}

\bibitem{akrout2016parallel}
A.~Akrout, A.~Bouwens, F.~Duport, Q.~Vinckier, M.~Haelterman, S.~Massar,
  arXiv:1612.08606  (2016)

\bibitem{vandoorne2014experimental}
K.~Vandoorne, P.~Mechet, T.V. Vaerenbergh, M.~Fiers, G.~Morthier,
  D.~Verstraeten, B.~Schrauwen, J.~Dambre, P.~Bienstman, Nature Communications
  \textbf{5}(1) (2014).
\newblock \doi{10.1038/ncomms4541}

\bibitem{triefenbach2010phoneme}
F.~Triefenbach, A.~Jalalvand, B.~Schrauwen, J.P. Martens, in \emph{Advances in
  neural information processing systems} (2010), pp. 2307--2315

\bibitem{NFC}
The 2006/07 forecasting competition for neural networks \& computational
  intelligence.
\newblock \url{http://www.neural-forecasting-competition.com/NN3/} (2006)

\bibitem{coarer2018all}
F.D.L. Coarer, M.~Sciamanna, A.~Katumba, M.~Freiberger, J.~Dambre,
  P.~Bienstman, D.~Rontani, {IEEE} Journal of Selected Topics in Quantum
  Electronics \textbf{24}(6), 1 (2018).
\newblock \doi{10.1109/jstqe.2018.2836985}

\bibitem{bueno2018reinforcement}
J.~Bueno, S.~Maktoobi, L.~Froehly, I.~Fischer, M.~Jacquot, L.~Larger,
  D.~Brunner, Optica \textbf{5}(6), 756 (2018).
\newblock \doi{10.1364/optica.5.000756}

\bibitem{antonik2019large}
P.~Antonik, N.~Marsal, D.~Rontani, {IEEE} Journal of Selected Topics in Quantum
  Electronics \textbf{26}(1), 1 (2020).
\newblock \doi{10.1109/jstqe.2019.2924138}

\bibitem{antonik2019human}
P.~Antonik, N.~Marsal, D.~Brunner, D.~Rontani, Nature Machine Intelligence
  (2019).
\newblock \doi{10.1038/s42256-019-0110-8}

\bibitem{dong2020optical}
J.~Dong, M.~Rafayelyan, F.~Krzakala, S.~Gigan, {IEEE} Journal of Selected
  Topics in Quantum Electronics \textbf{26}(1), 1 (2020).
\newblock \doi{10.1109/jstqe.2019.2936281}

\bibitem{penkovsky2018efficient}
B.~Penkovsky, L.~Larger, D.~Brunner, Journal of Applied Physics
  \textbf{124}(16), 162101 (2018).
\newblock \doi{10.1063/1.5039826}.
\newblock \urlprefix\url{http://aip.scitation.org/doi/10.1063/1.5039826}

\bibitem{mockus1994application}
J.~Mockus, Journal of Global Optimization \textbf{4}(4), 347 (1994).
\newblock \doi{10.1007/bf01099263}

\bibitem{brochu2010tutorial}
E.~Brochu, V.M. Cora, N.~De~Freitas, arXiv preprint arXiv:1012.2599  (2010)

\bibitem{mockus2012bayesian}
J.~Mockus, \emph{Bayesian approach to global optimization: theory and
  applications}, vol.~37 (Springer Science \& Business Media, 2012)

\bibitem{frazier2018tutorial}
P.I. Frazier, arXiv preprint arXiv:1807.02811  (2018)

\bibitem{yperman2016bayesian}
J.~Yperman, T.~Becker, arXiv:1611.05193  (2016).
\newblock \urlprefix\url{http://arxiv.org/abs/1611.05193}

\bibitem{griffith2019forecasting}
A.~Griffith, A.~Pomerance, D.J. Gauthier, Chaos: An Interdisciplinary Journal
  of Nonlinear Science \textbf{29}(12), 123108 (2019).
\newblock \doi{10.1063/1.5120710}

\bibitem{cerina2019lightweight}
L.~Cerina, G.~Franco, M.D. Santambrogio, in \emph{Proceedings of ESANN} (2019)

\bibitem{rasmussen2006gaussian}
C.E. Rasmussen, C.K. Williams, \emph{Gaussian process for machine learning}
  (MIT press, 2006)

\bibitem{tikhonov1995numerical}
A.N. Tikhonov, A.~Goncharsky, V.~Stepanov, A.G. Yagola, \emph{Numerical methods
  for the solution of ill-posed problems}, vol. 328 (Springer Netherlands,
  1995)

\bibitem{antonik2017online}
P.~Antonik, F.~Duport, M.~Hermans, A.~Smerieri, M.~Haelterman, S.~Massar,
  {IEEE} Transactions on Neural Networks and Learning Systems \textbf{28}(11),
  2686 (2017).
\newblock \doi{10.1109/tnnls.2016.2598655}

\bibitem{rodan2011minimum}
A.~Rodan, P.~Tino, {IEEE} Transactions on Neural Networks \textbf{22}(1), 131
  (2011).
\newblock \doi{10.1109/tnn.2010.2089641}

\bibitem{sande2017advances}
G.V. der Sande, D.~Brunner, M.C. Soriano, Nanophotonics \textbf{6}(3) (2017).
\newblock \doi{10.1515/nanoph-2016-0132}

\bibitem{mackay1992practical}
D.J.C. MacKay, Neural Computation \textbf{4}(3), 448 (1992).
\newblock \doi{10.1162/neco.1992.4.3.448}

\bibitem{matlab_fitrgp}
MathWorks.
\newblock Gaussian process regression model.
\newblock \url{http://fr.mathworks.com/help/stats/fitrgp.html}

\bibitem{matlab_bayes}
MathWorks.
\newblock Bayesian optimization algorithm.
\newblock
  \url{http://fr.mathworks.com/help/stats/bayesian-optimization-algorithm.html}

\bibitem{schuldt2004recognizing}
C.~Schuldt, I.~Laptev, B.~Caputo, in \emph{Proceedings of the 17th
  International Conference on Pattern Recognition, 2004. {ICPR} 2004.} ({IEEE},
  2004).
\newblock \doi{10.1109/icpr.2004.1334462}

\bibitem{dalal2005histograms}
N.~Dalal, B.~Triggs, in \emph{2005 {IEEE} Computer Society Conference on
  Computer Vision and Pattern Recognition ({CVPR})} ({IEEE}, 2005).
\newblock \doi{10.1109/cvpr.2005.177}

\bibitem{bahi2015robust}
H.E. Bahi, Z.~Mahani, A.~Zatni, S.~Saoud, {A robust system for printed and
  handwritten character recognition of images obtained by camera phone}.
\newblock Tech. rep. (2015).
\newblock
  \urlprefix\url{http://www.wseas.org/multimedia/journals/signal/2015/a045714-403.pdf}

\bibitem{pearson1901liii.}
K.~Pearson, The London, Edinburgh, and Dublin Philosophical Magazine and
  Journal of Science \textbf{2}(11), 559 (1901).
\newblock \doi{10.1080/14786440109462720}

\bibitem{hotelling1933analysis}
H.~Hotelling, Journal of Educational Psychology \textbf{24}(6), 417 (1933).
\newblock \doi{10.1037/h0071325}

\bibitem{smith2002tutorial}
L.I. Smith, A tutorial on principal components analysis.
\newblock Tech. rep. (2002)

\bibitem{shi2015learning}
Y.~Shi, W.~Zeng, T.~Huang, Y.~Wang, in \emph{2015 {IEEE} International
  Conference on Multimedia and Expo ({ICME})} ({IEEE}, 2015).
\newblock \doi{10.1109/icme.2015.7177461}

\end{thebibliography}

\end{document}